\documentclass{article}
\usepackage{amsmath}
\usepackage{amsfonts}
\usepackage[margin=1in]{geometry}
\usepackage{xcolor}
\usepackage{graphicx}
\usepackage{authblk}

\usepackage[backend=biber,style=numeric,sorting=none]{biblatex}
\title{IP-Basis PINNs: Efficient Multi-Query Inverse Parameter Estimation}
\author{Shalev Manor and Mohammad Kohandel \\
Department of Applied Mathematics, University of Waterloo, \\
Waterloo, Ontario, Canada, N2L 3G1}
\date{}

\begin{document}

\maketitle

\begin{abstract}
Solving inverse problems with Physics-Informed Neural Networks (PINNs) is computationally expensive for multi-query scenarios, as each new set of observed data requires a new, expensive training procedure. We present Inverse-Parameter Basis PINNs (IP-Basis PINNs), a meta-learning framework that extends the foundational work of Desai et al. (2022) to enable rapid and efficient inference for inverse problems. Our method employs an offline-online decomposition: a deep network is first trained offline to produce a rich set of basis functions that span the solution space of a parametric differential equation. For each new inverse problem online, this network is frozen, and solutions and parameters are inferred by training only a lightweight linear output layer against observed data. Key innovations that make our approach effective for inverse problems include: (1) a novel online loss formulation for simultaneous solution reconstruction and parameter identification, (2) a significant reduction in computational overhead via forward-mode automatic differentiation for PDE loss evaluation, and (3) a non-trivial validation and early-stopping mechanism for robust offline training. We demonstrate the efficacy of IP-Basis PINNs on three diverse benchmarks, including an extension to universal PINNs for unknown functional terms—showing consistent performance across constant and functional parameter estimation, a significant speedup per query over standard PINNs, and robust operation with scarce and noisy data.
\end{abstract}

\section{Introduction}
\label{introduction}
Physics-Informed Neural Networks (PINNs) have emerged as a transformative approach for solving differential equations (DEs) by integrating physical laws directly into the neural network's loss function. Unlike traditional numerical solvers that discretize the problem domain, PINNs leverage deep learning to define solutions over continuous spatial-temporal domains, making them particularly appealing for problems involving complex physical systems and parametric variations. Despite their promise, challenges remain, including slow training times and significant parameterization, which limit their efficiency in multi-query and real-time simulation settings.

Recent advancements, such as Generative Pre-Trained PINNs (GPT-PINNs), offer a meta-learning framework for parametric PDEs that addresses these issues by hyper-reducing the network's size while maintaining high accuracy across parameter domains \cite{chen_gpt-pinn_2024}. GPT-PINNs utilize pre-trained PINNs as activation functions within a meta-network, enabling efficient generation of surrogate solutions with reduced computational overhead. This approach parallels traditional model order reduction techniques like the Reduced Basis Method (RBM), which employs offline-online decomposition to achieve rapid computation during the online phase. The hybridization of PINNs with reduced basis concepts marks a significant step forward in bridging classical numerical methods with modern machine learning.

Another innovative development involves advanced PINN methodologies, such as $H^1$ loss functions and parallel PINNs (P-PINNs), which enhance training efficiency and accuracy \cite{ZHANG2024108229}. These methods address limitations in first-generation PINNs, such as limited accuracy with extensive training data, by introducing customized loss functions and specialized architectures to better handle coupled PDE systems. Similarly, parameterized neural network frameworks have been proposed for fiber transmission models, leveraging reduced basis expansion and parameterized coefficients for universal solution generation \cite{zang_fiber_2025}. These advancements underline the increasing applicability of PINNs in solving parametric systems efficiently and accurately.

Parallel to these developments, transfer learning has emerged as a viable set of methods to utilize pre-training in a set of tasks to assist training for unseen tasks\cite{zhuang_comprehensive_2020}. In a foundational work, \cite{desai_one-shot_2022} introduced a meta-learning framework for parametric PDEs using a network with multiple linear readouts and an offline-online decomposition. However, their work exclusively focused on the forward problem.

In this work, we extend this framework to introduce Inverse-Parameter Basis PINNs (IP-Basis PINNs), a method designed for efficient multi-query inverse problems. Our findings contribute to the growing body of literature advocating for hybrid approaches that combine classical numerical rigor with the flexibility of deep learning frameworks \cite{Koumpanakis2024MetalearningLF}.

The core contributions of IP-Basis PINNs are: (1) a novel online loss formulation for simultaneous solution reconstruction and parameter identification, extending the framework to inverse problems; (2) the use of forward-mode automatic differentiation for the evaluation of the PDE loss, which provides a significant reduction in the computational overhead required for the method's initial offline investment; and (3) a non-trivial implementation of early stopping using a validation loss, which stabilizes offline training and helps avoid overfitting.

\section{Methods}
\label{Methods}

In scenarios where repeated solutions to a set of inverse problems are required, the standard Physics Informed Neural Network (PINN) approach encounters significant inefficiencies. For every single inverse problem, a neural network must be trained from scratch to minimize the data and PDE losses. For large sets of inverse problems with similar features, this process is computationally wasteful and often too slow for time-sensitive applications.

To address this, we build upon the framework introduced by \cite{desai_one-shot_2022}, which employs an offline-online decomposition. We extend this framework to inverse problems and refer to our overall methodology as IP-Basis PINNs. The method performs an offline-online decomposition, allowing for an initial investment of computational resources in exchange for fast inference on a set of inverse problems. Sections \ref{Offline Method} and \ref{Online Method} describe the core offline-online mechanics from \cite{desai_one-shot_2022}. The subsequent sections detail our original contributions that enable and enhance this framework for inverse problems.

\subsection{Offline Phase}
\label{Offline Method}
In the offline phase, a large network with many outputs is constructed. We split this network into two components. The Network $R$ will be used to refer to the main body of the network, consisting of the input and hidden layers, and $L$ will refer to the linear output layer throughout the rest of this work. Thus, the network as a whole can be written as $NN(x) = (L \circ R) (x)$

Throughout this work, we consider the output from the network to be a collection of "readouts". Each readout is treated as the output of a separate PINN which may be composed of multiple network outputs. Each of these PINNs enforce the same PDE loss with different parameter values representative of the set of parameters expected to be encountered during online training. The loss of each individual readout can be expressed as \(\mathcal{L}_{PINN}(p_i,NN(\cdot)_i) = \omega_{PDE} \mathcal{L}_{PDE} +  \omega_{IC} \mathcal{L}_{IC} + \omega_{BC} \mathcal{L}_{BC}\) with
\[
\mathcal{L}_{PDE} = \frac{1}{N_{\mathrm{PDE}}} \sum^{N_{\mathrm{PDE}}}_{j=1}\big(Res(NN(x_j,t_j,\theta)_i,p_i)\big)^2
\]
\[
\mathcal{L}_{IC} = \frac{1}{N_{IC}} \sum^{N_{IC}}_{j=1}\big(NN(x_j,0,\theta)_i - u(x_j,0,p_i)\big)^2
\]
\[
\text{and } \mathcal{L}_{BC} = \frac{1}{N_{BC}} \sum^{N_{BC}}_{j=1}\big(NN(x_j,t_j,\theta)_i - u(x_j,t_j,p_i)\big)^2
\]
where \(i \in \{1,2,\dots,n\}\), n is the number of readouts, \(\mathcal{L}_{PDE}, \mathcal{L}_{IC},\) and \(\mathcal{L}_{BC}\) are the loss terms associated with conformity to the PDE, initial conditions, and boundary conditions, respectively. \(\omega_{PDE}, \omega_{IC},\) and \(\omega_{BC}\) are the weightings of the different loss terms. \((x_j,t_j)\) are sample points that may be different for each loss term. \(\theta\) represents the parameters of the network. $N_{\mathrm{PDE}},N_{IC},$ and $N_{BC}$ are the number of evaluation points for each term (evaluation points of $\mathcal{L}_{PDE}$ are commonly referred to as collocation points). Note that all evaluation points in $\mathcal{L}_{BC}$ are on the boundary. $u(x,t,p)$ is the solution to the parametric PDE problem with variables $x$ for position, $t$ for time, and $p$ for parameters. $Res(\cdot)$ is a differential operator that represents the residual of the PDE. To assist the network in optimizing towards a non-trivial solution, an additional loss term may be introduced to enforce conformance to collected data points on solutions to the PDE.

The network is trained through gradient descent to minimize the average loss over all of these PINNs, giving an overall loss of
\[
\overline{\mathcal{L}} = \frac{1}{n} \sum_{i=1}^n \mathcal{L}_{PINN}(p_i,(L \circ R)(\cdot)_i)
\]
where \(p_i\) is the \(i\)th set of parameters and \((L \circ R)(\cdot)_i\) is the \(i\)th readout of the neural network.
We observe that if the assumption is made that the final layer is optimized to its global minimum, the loss can be expressed as:
\[
\overline{\mathcal{L}} = \frac{1}{n} \sum_{i=1}^n \inf_{T} \mathcal{L}_{PINN}(p_i,(T \circ R)(\cdot))
\]
where T is a linear transformation of the form \(T:\mathbb{R}^w \to \mathbb{R}^d\). Here, \(w\) is the width of the final layer of \(R\) and \(d\) is the dimension of solutions to the PDE.
We also now see that this loss is the empirical approximation of the loss:
\[
\mathcal{L} = \mathbb{E}_{p \sim D}(\inf_{T} \mathcal{L}_{PINN}(p,T \circ R))
\]
Here $D$ is some distribution for the parameters. Finally, we observe that we are essentially treating $R$ as a basis for the solution space of the parametric PDE and optimizing it to represent solutions with parameters in the distribution $D$.

\subsection{Online Phase}
\label{Online Method}
For the online training component, a network $R$ trained in the offline component is combined with a new readout layer $L'$. In the inverse problem setting addressed by IP-Basis PINNs, the network is then trained to fit given sets of data points and identify the parameters used to generate them while only optimizing the parameters in $L'$ and the estimations of these generating parameters. 

The loss function used to achieve this is similar to the offline phase with a few key changes. First, a data loss term is used to enforce conformance to the given data. This term takes the form
\[
\mathcal{L}_{data} = \frac{1}{N_d} \sum^{N_d}_{j=1}\big((L' \circ R)(x_j,t_j,\theta)_i - u(x_j,t_j,p_i)\big)^2, 
\]
where \(u(x_j,t_j,p_i)\) are the given data points of the PDE with parameters \(p_i\). In many cases, even if the parameters are unknown, this term can be sampled by taking measurements of a physical system. The residual loss is also modified to account for our lack of knowledge of the parameters present in the data. This is done by replacing \(p_i\) with an estimate \(\bar p_i\), yielding
\[
\mathcal{L}_{PDE} = \frac{1}{N_c} \sum^{N_c}_{j=1}\big(Res((L' \circ R)(x_j,t_j,\theta)_i,\bar p_i)\big)^2
\]
In the IP-Basis PINN framework, these estimates are treated as trainable parameters and optimized along with the readouts during gradient descent. Through this process, a solution to the parametric PDE is found that matches the data along with estimates for the parameters.

A key insight is that the network $R$ is frozen during online training. Therefore, its outputs and derivatives at all collocation points can be precomputed and stored once, before online training begins. This avoids the need for expensive backpropagation through $R$ during each online training step. The required derivatives can be quickly calculated by applying the linear transformation of the readout layer to the stored derivatives of $R$, without the bias term since it goes to zero with the derivative.


With this method, the only operations that are required to calculate the PINN loss at each training step are a few matrix multiplications and arithmetic, instead of full backpropagation through the entire network which requires many matrix multiplications. This allows for significant speedups when comparing the offline training to the online component which we demonstrate in Section \ref{results}. This precomputation strategy, introduced by \cite{desai_one-shot_2022} for the forward problem, is a key enabler of efficiency.

IP-Basis PINNs allow for an initial investment of computational resources in exchange for fast inference on a set of inverse problems. The method stands to reduce computational resource requirements and inference time over the standard PINN approach for applications which require repeated solutions to similar inverse problems.

\subsection{Forward-Mode Auto-Differentiation For Calculating PINN Loss}
\label{AutoDiff}

Auto-Differentiation is an integral component in the majority of modern implementations of neural networks. The specific methods of implementation are often not explicitly considered. The term "Auto-Differentiation" refers to a family of methods as opposed to a single algorithm \cite{baydin_automatic_2018}. Each method comes with its own sets of advantages and disadvantages. Backward-mode auto-differentiation, commonly referred to as backpropagation is the primary method used by PyTorch \cite{NEURIPS2019_bdbca288} and TensorFlow \cite{tensorflow2015-whitepaper}. Backward mode excels at calculating the gradients of functions with a large number of inputs and small number of outputs. This makes it an excellent choice for use in the gradient descent optimization of neural networks on loss functions. This is because losses are functions of all network parameters with a single output, which aligns perfectly with the strengths of backward mode.

In the context of PINNs, auto-differentiation is used in both the calculation and optimization of the loss. To calculate the loss, the derivative of the network output is required with respect to the input. In this context, backward mode performs considerably worse than for optimization. There are two reasons for this: First, the derivatives of each network output must be taken individually with a separate call to the algorithm. This is particularly costly in the IP-Basis PINN framework, where the network $R$ has many outputs (one for each basis function), each requiring its own derivative calculation. And second, to compute second derivatives (which are commonly required to evaluate the PINN loss) using backward mode requires calling the algorithm on its own output from the first derivative. Since the process of backpropagation itself grows the computational graph (which is required to calculate the result), this second call is even more expensive to compute than the first. For standard PINNs on low dimensional PDEs this inefficiency is not severe but as the number of outputs scales (as is required in the IP-Basis PINN framework) it becomes more pronounced. This motivates the use of an alternative approach to auto-differentiation for the calculation of the PINN losses.

\subsubsection{Forward Mode Auto Differentiation}

Forward mode auto differentiation is based on Taylor series expansions. We examine a truncated Taylor series of a function \(f\) centered at \(\vec x\) evaluated at \(\vec x + \vec d\) where \(\vec d\) is some disturbance:
\[
f(\vec x + \vec d) = f(\vec x) + \nabla f(\vec x)\vec d  + \frac{1}{2}\vec d^T \nabla^2f(\vec x)\vec d
\]
Where \(\vec d^T\) is the transpose of \(\vec d\) and \(\nabla^2f(\vec x)\) is the Hessian matrix of \(f\) evaluated at \(\vec x\). We now replace the disturbance \(\vec d\) with a set of variables \(\epsilon_1,\epsilon_2\) and \(\epsilon_1 \epsilon_2\) each multiplied by corresponding vectors commonly referred to as the "tangent vectors" \(\vec v_1,\vec v_2\) and \(\vec v_{12}\). We take that \(\epsilon_1,\epsilon_2,\epsilon_1 \epsilon_2 \neq 0\) but are sufficiently small to make the approximation \(\epsilon_1^2 = \epsilon_2^2 = (\epsilon_1 \epsilon_2)^2 = 0\) which is used when calculating the output of \(f\). This yields \(\vec d = \vec v_1 \epsilon_1 +\vec v_2 \epsilon_2 + \vec v_{12} \epsilon_1 \epsilon_2\). Plugging this back into the Taylor series yields:
\[
f(\vec x + \vec v_1 \epsilon_1 +\vec v_2 \epsilon_2 + \vec v_{12} \epsilon_1 \epsilon_2) = f(\vec x) + \nabla f(\vec x)(\vec v_1 \epsilon_1 +\vec v_2 \epsilon_2 + \vec v_{12} \epsilon_1 \epsilon_2) + \vec v_1^T \nabla^2f(\vec x)\vec v_2\epsilon_1 \epsilon_2
\]
This implies that if we plug \(\vec x + \vec v_1 \epsilon_1 +\vec v_2 \epsilon_2 + \vec v_{12} \epsilon_1 \epsilon_2\), which is referred to as a hyper-dual vector, into our function \(f\), the result will also be a hyper-dual vector. The component that is not multiplied by any epsilons will be equivalent to the output of the function at \(\vec x\), the components that are multiplied by \(\epsilon_1\), \(\epsilon_2\), and \(\epsilon_1\epsilon_2\) are the dot products between the gradient of the function at \(\vec x\) and the tangent vectors \(\vec v_1\), \(\vec v_2\), and \(\vec v_{12}\) respectively, and the final component multiplied by \(\epsilon_1\epsilon_2\) corresponds to a linear combination of the entries of the Hessian matrix as specified by \(\vec v_1\) and \(\vec v_2\). 

To sample from the gradient and Hessian, \(\vec v_1\) and \(\vec v_2\) are set to \(\vec e_i\) and \(\vec e_j\) respectively, which are vectors with ones in rows \(i\) and \(j\) respectively and zeros everywhere else. \(\vec v_{12}\) is set to zero. The output of the function when given this hyper-dual is
\[
f(\vec x + \vec e_i \epsilon_1 +\vec e_j \epsilon_2) = f(\vec x) + \nabla f(\vec x)(\vec e_i \epsilon_1 +\vec e_j \epsilon_2) + \vec e_i^T \nabla^2f(\vec x)\vec e_j\epsilon_1 \epsilon_2
\]
which gives us the network output, the \(i\)th and \(j\)th components of the gradient, and the entry of the Hessian in row \(i\) and column \(j\). These first and second derivatives can then be used to calculate the PINN losses of the readout layer. A key advantage in the IP-Basis PINN context is that, for a given input point, the derivatives of all outputs of network $R$ with respect to that input can be computed simultaneously in a single forward pass. The network loss can then be optimized using backward mode auto differentiation.

The current distribution of PyTorch does not have an implementation of hyper-dual vectors and is thus unable to calculate second order derivatives without backpropagation. Fortunately, thanks to the work in \cite{cobb_second-order_2024} and its corresponding open source repository: Fomoh, we were able to quickly implement forward mode auto differentiation for the calculation of the loss function. The work in \cite{cobb_second-order_2024} was also a large influence in the writing and presentation of this section. For more details on the methods described above, please see their work and the references within.

\subsection{Validation And Early Stopping}
\label{Validation}

For many machine learning tasks, it is common to implement early stopping based on model performance on a validation set that is not used to calculate gradients for training. It is common to see performance on the validation set improve alongside the training set for some time. This continues until the validation loss reaches a minimum and begins to rise while the training loss continues to decrease. Performance on the validation set is taken as a proxy of the model's expected loss, and by definition the training loss is the empirical loss. Since we would like to minimize the expected loss of the network, the model with the lowest validation loss is taken as the final result of the training algorithm. Thus, the training procedure can be stopped once the validation loss reaches its minimum. In practice, this is done by breaking out of the training loop after a set number of epochs without any improvement on the validation loss, returning the model which achieved the lowest loss.

Implementing a validation set that is queried inside the offline training loop to allow for model check-pointing and early stopping is non-trivial for the IP-Basis PINN framework. In this section, we outline our approach to solving this.

Our goal is to minimize the expected loss
\[
\mathcal{L} = \mathbb{E}_{p \sim D}(\inf_{T} \mathcal{L}_{PINN}(p,T \circ R))
\]
described in Section \ref{Offline Method}. This is done by performing gradient descent on the networks \(R\) and \(L\) to minimize the empirical loss
\[
\overline{\mathcal{L}} = \frac{1}{n} \sum_{i=1}^n \mathcal{L}_{PINN}(p_i,(L \circ R)(\cdot)_i)
\]
which corresponds to optimizing the performance of the neural network \((L \circ R)\) to perform inference on \(n\) PINNs of differing parameters \(p_i\) simultaneously. We now define the validation loss as
\[
\tilde{\mathcal{L}} = \frac{1}{m} \sum_{i=1}^m \mathcal{L}_{PINN}(\tilde p_i,(\tilde L \circ R)(\cdot)_i)
\]
where \(\tilde p_i\) are new parameters drawn from the same distribution as the parameters \(p_i\), \(\tilde L\) is a new readout layer with \(m\) readouts. For this loss to be used for model check-pointing, we need it to approximate the best loss that would be achievable using the network \(R\) during the online training phase. Thus, ideally, \(\tilde L\) would be fully trained to minimize the validation loss at each epoch of offline training. In practice, this is quite impractical since it would significantly increase the training time of the offline phase. Instead, we take advantage of the fact that the gradient descent optimization process makes small changes between steps and optimize \(\tilde L\)  while taking only one step per epoch. Thus in each epoch of the offline training, the network \((L \circ R)\) takes a step to optimize the training loss and \(\tilde L\) takes a step to optimize the validation loss.

If the forward mode auto-differentiation  described in Section \ref{AutoDiff} is used, optimizing the validation loss would not incur a significant cost in resources. This is because the residual loss can be calculated using the reused derivatives of the final layer of \(R\) from the calculation of the training loss with only a few matrix multiplications. Since the loss only depends on \(\tilde L\), backpropagation only needs to calculate the gradients with respect to a single linear transformation similar to the online phase, making this calculation relatively lightweight.

\section{Results}
\label{results}

To validate the IP-Basis PINN framework, we conducted three sets of experiments, solving a range of inverse problems. We first used the method to estimate the parameters of a damped harmonic oscillator from data. Next, we applied it to find the predator-prey interaction terms in the Lotka-Volterra system of differential equations using Universal PINNs \cite{podina_universal_2023}. Lastly, we tested the model on the time-dependent Schr\"odinger equation, estimating the force constant of the quantum harmonic oscillator.

\subsection{Damped Harmonic Oscillator}
\label{DHO}
The first experiment we conducted was on the damped harmonic oscillator, the solutions of which were defined to follow the ordinary differential equation,
\begin{equation}
    \frac{d^2x}{dt^2} + \alpha \frac{dx}{dt} + \beta x = f
\end{equation}
where \(\alpha\), \(\beta\), and \(f\) are real valued constants. In our experiment, we implemented Basis Networks to estimate the values of \(\alpha\) and \(\beta\) given a set of points using the methods described in Section \ref{Methods}.

For the offline phase, the network \(R\) had a single input for time, 4 hidden layers with widths of 40, and tanh activations. For the readout layer, we tested networks trained with 10, 30, and 50 readouts and stored the trained networks. We will refer to the networks as \(R_{10}\), \(R_{30}\), and \(R_{50}\) respectively. Each of the readouts were randomly assigned values for \(\alpha\), \(\beta\), and \(f\) and the individual losses used these parameters. The values for \(\alpha\) and \(\beta\) were drawn uniformly between 0 and 1.5 and \(f\) between -1.5 and 1.5. The initial conditions were drawn uniformly between -5 and 5 for both position and velocity. The time span for the trajectories was set to three non-dimensionalized time units with 30 evenly spaced collocation points. Training was conducted with a learning rate of \(5 \times 10^{-5}\) and was run for 30000 epochs. Throughout the offline training a validation loss was utilized according to the methods described in Section \ref{Validation} with 100 validation models with parameters and initial conditions drawn from identical distributions. The validation readout layer was trained with a learning rate of \(3 \times 10^{-2}\). Both the offline and online training routines were run on CPU clusters with access to two Intel Xeon Gold 6244 8-core 3.6 GHz (Cascade Lake). It is important to note that the exact training times varied per trial to a large extent due to variable loads on the shared cluster. Therefore, the absolute times reported in Tables \ref{DHO offline} and \ref{DHO online} should be interpreted as approximate. However, the relative comparisons, such as the significant speedup between forward and backward mode autodiff, are robust as they were computed on the same hardware under similar conditions. The results of the offline phase are listed in Table \ref{DHO offline}.

For the online phase, a pre-trained network that is not included in those discussed in this work was paired with a new readout layer which had 10 individual readouts corresponding to different solutions to the ODE. These readouts were treated as a validation set and used to tune the hyperparameters for online training. We make the assumption that for this set the true values of the parameters were known and thus tuned the hyperparameters to minimize the mean absolute and squared errors. Any of the networks discussed in this section would suffice to use but after modifying our experimental procedures we found that the same hyperparameters worked for the online phase so refrained from retuning them.

The hyperparameters optimized using the validation set were then used to perform online inference. Parameters were drawn from the same distributions as in the offline training and the initial conditions were drawn from uniform distributions with ranges from -5 to 5 as in the offline phase up to -40 to 40. These were used as inputs to an ODE solver to generate the corresponding solutions, yielding 100 data points evenly spaced over time which were fed into the data loss. The random seed was changed between the validation and test set evaluations. The collocation points were identical to the offline phase. The performance of online inference using \(R_{10}\),\(R_{30}\), and \(R_{50}\) was evaluated. The learning rate used was \(5 \times 10 ^{-2}\) for the first 5000 epochs and \(5 \times 10 ^{-3}\) for the rest of the 15000 total. The loss weights were set to \(\omega_{Data} = 1\) and \(\omega_{ODE} = 0.001\). The rest of the weights were set to zero, as those terms are not required for inverse problems on ODEs. The results of the online phase are shown in Table \ref{DHO online}. 

\begin{table}[h!]
\centering
\begin{tabular}{| c | c | c | c | c | c | c |} 
 \hline
  & Readouts & Final Loss & Final Val. Loss & Val. Pred. Param. MSE & Training Time (Seconds) \\ [0.5ex] 
 \hline\hline
 \(R_{10}\) & 10 & 1.286E-6 & 9.336E-4 & 2.170E-2 & 193.648 \\ 
 \hline
 \(R_{30}\) & 30 & 1.804E-6 & 1.184E-3 & 1.179E-2 & 191.175 \\
 \hline
 \(R_{50}\) & 50 & 2.478E-6 & 2.373E-3 & 2.131E-2 & 201.950 \\
 \hline
\end{tabular}
\caption{Damped Harmonic Oscillator offline training results.}
\label{DHO offline}
\end{table}

\begin{table}[h!]
\centering
\begin{tabular}{| c | c | c | c | c | c | c | c |} 
 \hline
  & IC \(\in\) & Final Loss & TT (Sec) & \(P_1\) MSE & \(P_1\) MAE & \(P_2\) MSE & \(P_2\) MAE \\ [0.5ex] 
 \hline\hline
 \(R_{10}\) & \([-5,5]\) & 1.377E-5 & 29.460 & 3.211E-2 & 2.902E-2 & 1.548E-3 & 1.052E-2 \\ 
 \hline
 \(R_{10}\) & \([-10,10]\) & 1.558E-5 & 29.278 & 3.437E-2 & 3.155E-2 & 5.208E-3 & 1.521E-2 \\ 
 \hline
 \(R_{10}\) & \([-20,20]\) & 2.254E-5 & 29.857 & 3.376E-1 & 9.197E-2 & 1.383E-1 & 5.741E-2 \\
 \hline
 \(R_{10}\) & \([-40,40]\) & 4.713E-5 & 28.961 & 1.077 & 1.748E-1 & 5.607E-1 & 1.218E-1 \\
 \hline\hline
 \(R_{30}\) & \([-5,5]\) & 1.724E-5 & 29.105 & 6.656E-2 & 3.898E-2 & 2.556E-3 & 1.284E-2 \\ 
 \hline
 \(R_{30}\) & \([-10,10]\) & 1.569E-5 & 29.278 & 6.582E-2 & 4.321E-2 & 8.289E-3 & 1.962E-2 \\ 
 \hline
 \(R_{30}\) & \([-20,20]\) & 1.804E-5 & 29.792 & 1.206E-1 & 6.711E-2 & 4.742E-2 & 4.284E-2 \\ 
 \hline
 \(R_{30}\) & \([-40,40]\) & 4.203E-5 & 29.163 & 2.658E-1 & 1.077E-1 & 1.506E-1 & 8.005E-2 \\ 
 \hline\hline
 \(R_{50}\) & \([-5,5]\) & 1.604E-5 & 29.514 & 1.688E-1 & 5.248E-2 & 3.573E-3 & 1.357E-2 \\ 
 \hline
 \(R_{50}\) & \([-10,10]\) & 1.529E-5 & 29.677 & 2.031E-1 & 6.650E-2 & 1.909E-2 & 2.581E-2 \\ 
 \hline
 \(R_{50}\) & \([-20,20]\) & 1.831E-5 & 29.496 & 2.838E-1 & 9.548E-2 & 8.745E-2 & 5.639E-2 \\ 
 \hline
 \(R_{50}\) & \([-40,40]\) & 5.671E-5 & 30.630 & 3.862E-1 & 1.298E-1 & 2.048E-1 & 9.459E-2 \\
 \hline
\end{tabular}
\caption{Damped Harmonic Oscillator online training results. "TT" corresponds to "Training Time". "\(P_1\)" and "\(P_2\)" refer to the two parameters on which the model performs inference}
\label{DHO online}
\end{table}

We observe that the method was able to successfully infer the parameters from the given data with some accuracy. Surprisingly, it seems that the network that performed best on the initial distribution of initial conditions was \(R_{10}\). On the other hand, \(R_{30}\) and \(R_{50}\) performed better on a wider set of initial conditions. While the performance of \(R_{30}\) and \(R_{50}\) is quite similar, it seems that \(R_{30}\) maintained an edge over all trials.

To test the effect of forward mode auto differentiation on training speed, we reran the offline training procedure for \(R_{50}\) using backward mode on the same cluster. Despite potential variability, the training loop took 1020.971 seconds to conclude. Compared to this value, the forward mode time of 176.091 seconds represents a speedup of approximately 5.8x, highlighting the substantial efficiency gain of our approach.

\begin{figure}
    \centering
    \includegraphics[width=0.5\linewidth]{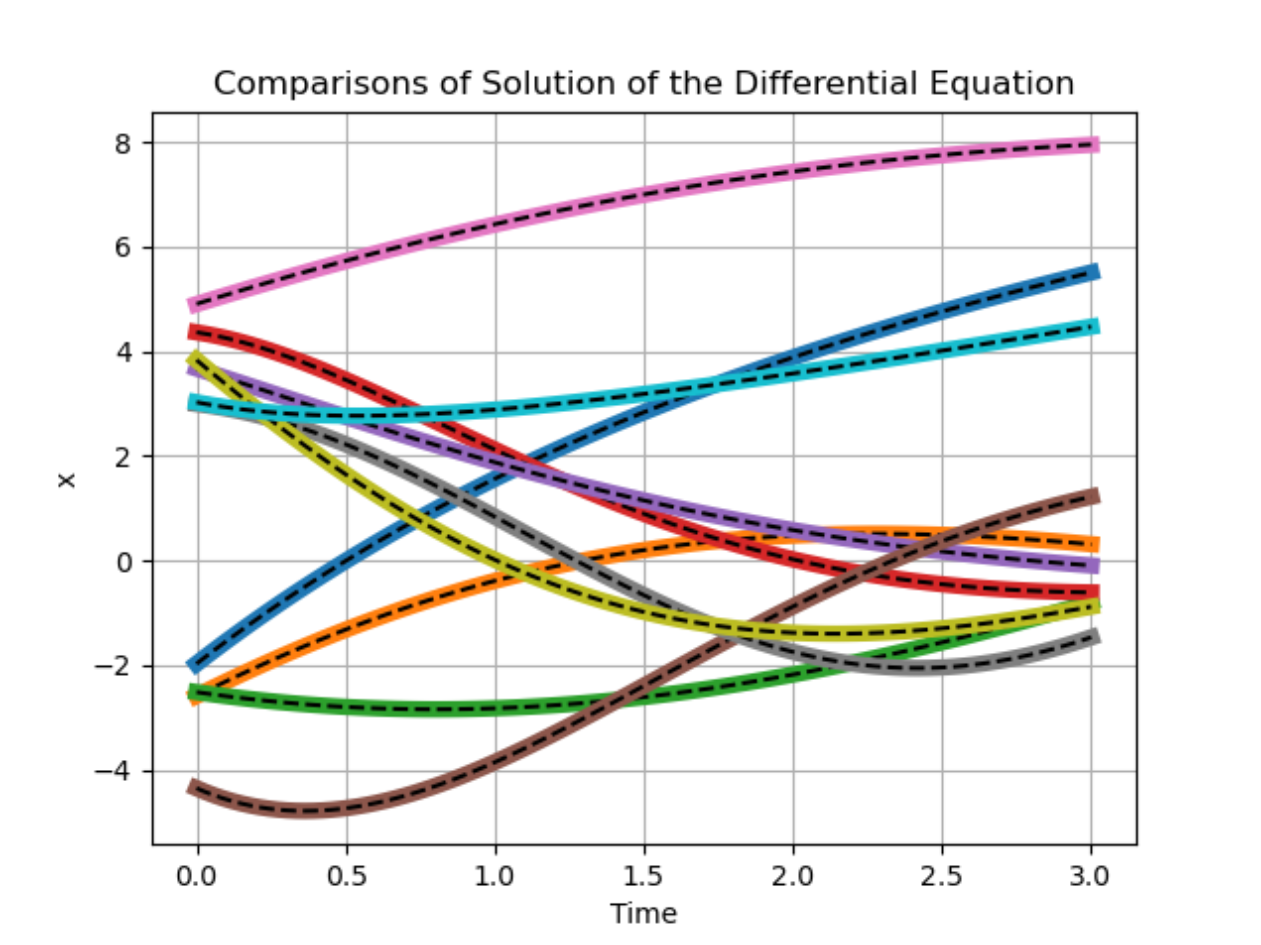}
    \caption{Comparison between numerical solutions (dashed lines) and online model approximations (solid lines) for randomly generated parameters of the Damped Harmonic Oscillator. Presentation inspired by \cite{desai_one-shot_2022}.}
    \label{fig:DHOResults}
\end{figure}

A plot of 10 readouts trained online using \(R_{50}\) and the procedure described above is shown in Figure \ref{fig:DHOResults}.

\subsection{Lotka-Volterra Ecological Model}
UPINNs extend the range of inverse problems that can be tackled by PINNs. Where PINNs can find unknown multiplicative constants in an equation with known functional terms, UPINNs find entirely unknown terms of the differential equation. This is done by replacing the unknown components in the system with neural networks. This new system is then used in the residual loss and the new networks are optimized similarly to the unknown parameters in PINNs. 

To illustrate the applicability of the IP-Basis PINN framework to UPINNs \cite{podina_universal_2023}, we examine the Lotka-Volterra ecological model which is described by the dynamical system
\begin{equation}
\label{LV1}
    \frac{dx}{dt} = \alpha x - \beta xy
\end{equation}
\begin{equation}
\label{LV2}
    \frac{dy}{dt} = -\gamma y + \delta xy
\end{equation}
where \(\alpha, \beta, \gamma,\) and \(\delta\) are nonnegative constants.

The offline phase remained unchanged from Section \ref{Methods}, where we simply train \(R\) to be a good basis for the solution space of the parametric ODE.
During the online phase, we treated \(\beta xy\) and \(\delta xy\) from equations \ref{LV1} and \ref{LV2} as unknown terms. These terms were replaced with neural networks which were optimized along with the main network in order to approximate the unknown terms according to the methods described in \cite{podina_universal_2023}.


During the offline phase, \(R\) had a single input \(t\) for time, 4 hidden layers of width 64, and tanh activations. \(L\) took in the 64 inputs from \(R\) and mapped them to differing numbers of readouts. We tested the performance of networks trained offline with 10, 30, 50, and 100 readouts, referred to as \(R_{10},R_{30},R_{50},\) and \(R_{100}\), respectively. Each readout was randomly assigned a set of parameters \(\alpha, \beta, \gamma,\) and \(\delta\) all randomly drawn from a uniform distribution between 0.5 and 1.5 and a set of initial conditions in a square with vertices at the points \((0,0)\) and \((2,2)\). The time span for training was from \(t = 0\) to \(t = 10\) and 1000 collocation points were evenly spaced throughout. To circumvent issues that we encountered of the network finding and perturbing itself onto trivial solutions, we introduced a data loss term using points generated with an ODE solver on times matching the collocation points. The training loop was run for 40000 epochs with a learning rate of \(3 \times 10^{-4}\). We did not implement early stopping with a validation loss for this section, instead opting to use model check-pointing to get the network with the lowest training loss. In this section, both the offline and online training routines were again run on CPU clusters with access to two Intel Xeon Gold 6244 8-core 3.6 GHz (Cascade Lake). 
The results are listed in Table \ref{LV offline}.

\begin{table}[h!]
\centering
\begin{tabular}{| c | c | c |} 
 \hline
  & Final Loss &  Training Time (Seconds) \\ [0.5ex] 
 \hline\hline
 \(R_{10}\) & 1.631E-5 & 601.675 \\
 \hline
 \(R_{30}\) &  5.102E-5 & 692.726 \\
 \hline
 \(R_{50}\) &  4.750E-5 & 699.155 \\
 \hline
 \(R_{100}\) &  1.574E-4 &  859.105 \\
 \hline
 \(R_{100}\textsuperscript{1}\) &  4.697E-5 &  4944.725 \\
 \hline
\end{tabular}
\caption{Lotka-Volterra system offline training results. \textsuperscript{1} Utilized backward mode auto-differentiation.}
\label{LV offline}
\end{table}

The result for $R_{100}$ (Bkwrd Mode) clearly demonstrates the substantial efficiency gain achieved by using forward-mode autodiff, as proposed in our method. While absolute timings may vary with server load, the order-of-magnitude difference (859 vs. 4944) is a robust indicator of the performance improvement.


The online evaluation was again divided into two phases. Hyperparameter optimization in a small validation set, followed by model evaluation in a large test set. For the hyperparameter optimization, a pretrained network was paired with a new readout layer with 10 individual readouts. The parameters were chosen at random according to the same distribution as the offline phase. The initial conditions were drawn uniformly from within a square with vertices at \((0.1,0.1)\) and \((2.1,2.1)\). This was done to alleviate the effects of the instability of the system near the origin. We observed a significant improvement in test performance by shifting the initial condition distribution as described. The same number of data points was used as in the offline phase. The final parameters to be used on the validation and testing set were: a learning rate of \(3 \times 10^{-3}\) loss weights of \(\omega_{PDE} = 0.1\) and \(\omega_{Data} = 1\) all run for 10000 epochs. For the test set, a new final layer with 100 readouts was attached to the pretrained networks from the offline phase. Parameters and initial conditions for these readouts were randomly generated according to the same distributions as the validation set with a new random seed. The results are listed in Table \ref{LV online}. It is important to note that the parameter MSE and MAE for the unknown terms are not calculated in the same way as in Section \ref{DHO}. To obtain the values listed in the table, the neural networks representing the unknown terms were evaluated on the data points seen throughout training. The MSE and MAE of these values were then calculated by comparing to the output of the true unknown term output. The listed values are the average MSE and MAE over all readouts.

\begin{table}[h!]
\centering
\begin{tabular}{| c | c | c | c | c |} 
 \hline
  & Final Loss & Unknown Term MMSE & Unknown Term MMAE & Training Time (Seconds) \\ [0.5ex] 
 \hline\hline
 \(R_{10}\) & 1.770E-2 & 2.886E-1 & 8.553E-2 & 1063.626 \\
 \hline
 \(R_{30}\) & 5.927E-4 & 3.004E-2 & 3.088E-2 & 893.452 \\
 \hline
 \(R_{50}\) & 8.931E-4 & 4.893E-2 & 2.572E-2 & 906.864 \\
 \hline
 \(R_{100}\) & 4.315E-4 & 2.596E-2 & 2.409E-2 & 865.860 \\
 \hline
\end{tabular}
\caption{Lotka-Volterra system online training results. "MMSE" and "MMAE" refer to the mean values of the mean squared and absolute errors of the individual unknown term predictions respectively. Training times are reported for completeness but should be interpreted with the note on cluster variability in mind.}
\label{LV online}
\end{table}

Note that the reason why the online training in this section takes significantly longer is due to the number of parameters undergoing optimization. In sections \ref{DHO} and \ref{QHO}, only a linear transformation and a few more parameters for the inverse problem must be optimized. In this section, we optimized a linear transformation along with a small neural network for each readout.

We observe that the method was able to successfully recover a large number of unknown terms from the given data with high accuracy for all of the pre-trained networks. The accuracy of the method increased along with the number of readouts used in offline training but with diminishing returns. The increase from 10 to 30 offline training readouts more than doubled accuracy but from 50 to 100 readouts only yielded a slight improvement. The variations in online training time are likely entirely caused by variations in the CPU server as the code for all trials was identical.

Next, we tested the performance of the pre-trained networks for online inference on previously unseen parameter values. We ran this trial with an identical procedure except for the parameter distribution which was taken as uniform between 1.5 and 2.5 instead of the previous 0.5 to 1.5. The results for this trial are shown in Table \ref{LV online OOD}.

\begin{table}[h!]
\centering
\begin{tabular}{| c | c | c | c | c |} 
 \hline
  & Final Loss & Unknown Term MMSE & Unknown Term MMAE & Training Time (Seconds) \\ [0.5ex] 
 \hline\hline
 \(R_{10}\) & 2.911E-2 & 4.062E-1 & 2.827E-1 & 934.634 \\
 \hline
 \(R_{30}\) & 6.797E-3 & 2.606E-1 & 1.564E-1 & 885.282 \\
 \hline
 \(R_{50}\) & 3.141E-3 & 1.785E-1 & 1.108E-1 & 895.664 \\
 \hline
 \(R_{100}\) & 4.154E-3 & 2.357E-1 & 1.270E-1 & 884.664 \\
 \hline
\end{tabular}
\caption{Lotka-Volterra system online training results for parameters out of distribution. Training times are reported for completeness but should be interpreted with the note on cluster variability in mind.}
\label{LV online OOD}
\end{table}

Although the accuracy was significantly reduced for parameters outside the training distribution, the networks were still able to train online to a large extent. the same trends as in Table \ref{LV online} appear here. The networks trained with more readouts offline tended to perform better. This pattern is broken between \(R_{50}\) and \(R_{100}\) as all measures of performance list \(R_{100}\) as performing worse. This could have been caused by chance throughout the training process or be the result of overfitting to the training distribution. We plot 10 readouts trained online using \(R_{100}\) and the procedure described above in Figure \ref{fig:DHOResults}.

\begin{figure}[h]
    \centering
    \includegraphics[width=0.5\linewidth]{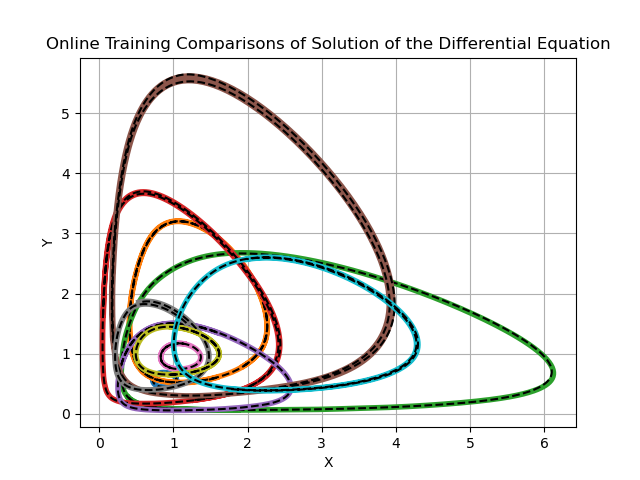}
    \caption{Comparison between numerical solutions (dashed lines) and online model approximations (solid lines) for randomly generated parameters of the Lotka-Volterra system. Presentation inspired by \cite{desai_one-shot_2022}}
    \label{fig:LVResults}
\end{figure}

\subsection{Quantum Harmonic Oscillator}
\label{QHO}

We now demonstrate IP-Basis PINNs' applicability to PDEs by using it to solve for the force constant in the one dimensional quantum harmonic oscillator.
The dimensionless quantum harmonic oscillator is described by the PDE
\[
i\frac{\partial}{\partial t}\Psi(x,t) = \Big[-\frac{1}{2}\frac{\partial^2}{\partial x^2} + V(x) \Big]\Psi(x,t)
\]
where \(\Psi(x,t)\) is the wave function and \(V(x)\) is the potential function. For the quantum harmonic oscillator, \(V(x) = \frac{1}{2}kx^2\) where k is some real "force constant". Since neural networks do not natively support complex numbers, we took a similar approach to \cite{raissi_physics-informed_2019} and \cite{desai_one-shot_2022} by splitting the output into real and imaginary components. Numerical solutions used to assist in offline training and provide data to test performance were generated using the finite element method implemented in FEniCSx \cite{BarattaEtal2023}. 

For the offline phase, the model was trained in a computational domain of \(x \in [-5,5], t \in [0,1.5]\) with collocation points in a 100 by 100 grid yielding a total of 10000. The initial conditions used throughout all phases were
\[u(x,0,p) = A \, \text{exp}\Big(\frac{-(x+2)^2}{2\sigma^2}\Big) \, \text{exp}(i \tilde v x)\]
Here \(u(x,0,p)\) is the PDE solution at \((x,t=0)\), for parameter \(p\), \(\sigma = 0.5\) is the width of the initial wave packet, \(\tilde v = 0.1\) is the wave number, and \(A = 1/(\sigma\sqrt\pi)\) is the normalization constant. The main body of the model, \(R\) had inputs for location and time, 5 hidden layers of width 100, and tanh activations. The readout layer contained 6 readouts with \(k = 0,1,\dots,5\). Each readout utilized 10000 data points from numerical solutions with \(k\)'s matching the readout's assigned value. All data point locations and times were randomly sampled but were consistent between readouts. The model was trained for 3200 epochs with a learning rate of \(2 \times 10^{-3}\) until the final 200 epochs which utilized a learning rate of \(2 \times 10^{-4}\). The weights of the different loss terms were set to \(\omega_{IC} = 1\) and \( \omega_{BC} = 1 \times 10^{-3}\). \(\omega_{PDE}\) was initialized with a value of \(5 \times 10^{-4}\) but is linearly interpolated to \(5 \times 10^{-2}\) between epochs 1000 and 3000. \(\omega_{Data}\) initialized with a value of \(1\) and was divided by 2 at epochs 1000, 1500, 2000, 2500, and 3000. The choice of weights seemed to affect the final outcome of training significantly and the final values were chosen such that the network would not converge to trivial solutions. In this section, the offline training routine was run on a GPU cluster with 2x AMD EPYC 7542 2.9 GHz with a single partitioned NVIDIA Ampere A100 PCI. Throughout training, the best loss achieved by the network was 9.010E-4. The training loop took 437.420 seconds to conclude.

The online phase was again split into a validation phase where hyperparameter were tuned and a testing phase, where the method performed inference on completely unseen parameters and the results were taken for evaluation. The online phase was again run on the CPU cluster.

In the validation phase, a new readout layer was trained from scratch on the same parameter values and data points used in the offline training (\(k = 0,1,\dots,5\)). The hyperparameters were tuned to minimize the mean absolute and squared errors, making the assumption that the \(k\) values were known. The final hyperparameters achieved through this process were as follows: 6000 training epochs, a learning rate of \(5 \times 10^{-3}\), which is brought down to \(2.5 \times 10^{-3}\) in epoch 5000, and loss function weights of \(\omega_{PDE} = 1.5 \times 10^{-2}, \omega_{IC} = 1 \times 10^{-2}, \omega_{BC} = 1 \times 10^{-2},\) and \(\omega_{Data} = 1\).

For the testing phase, the same hyperparameters were used to perform inference for values of \(k\) not observed during training. We chose to use \(k = \frac{1}{3},\frac{2}{3},\frac{4}{3},\frac{5}{3},\dots,\frac{14}{3}\), resulting in 10 readouts. We tested the accuracy of inference when given different sizes of training sets. Starting with 10,000 data points as used in offline training and validation, the set size was divided by 10 until a set of size 10 was reached. The results are shown in Table \ref{QHO online}.

\begin{table}[h!]
\centering
\begin{tabular}{| c | c | c | c | c | c |} 
 \hline
  & Num. Data Points & Final Loss & Param. MSE & Param. MAE & TT (Seconds) \\ [0.5ex] 
 \hline\hline
 Validation & 10,000 & 6.542E-04 & 7.010E-03 & 5.590E-02 & 130.856 \\ 
 \hline\hline
 Test & 10,000 & 4.793E-04 & 2.110E-03 & 3.189E-02 & 187.842 \\
 \hline
 Test & 1,000 & 4.839E-04 & 3.356E-03 & 3.728E-02 & 139.615 \\
 \hline
 Test & 100 & 5.276E-04 & 2.305E-02 & 1.031E-01 & 136.874 \\
 \hline
 Test & 10 & 6.817E-04 & 6.551E-01 & 6.499E-01 & 135.533 \\
 \hline
\end{tabular}
\caption{Quantum Harmonic Oscillator online training results. "Param." is used as shorthand for "Parameter" and "TT" for "Training Time".}
\label{QHO online}
\end{table}

\begin{figure}[h]
    \centering
    \includegraphics[width=0.6\linewidth]{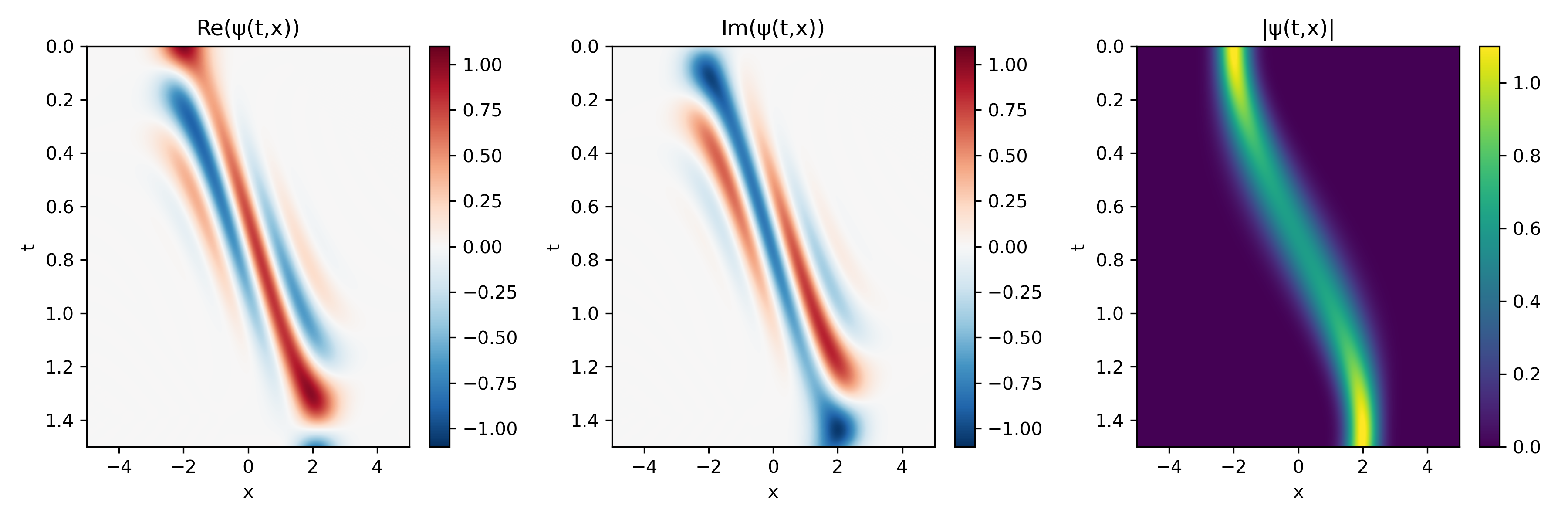}
    \includegraphics[width=0.6\linewidth]{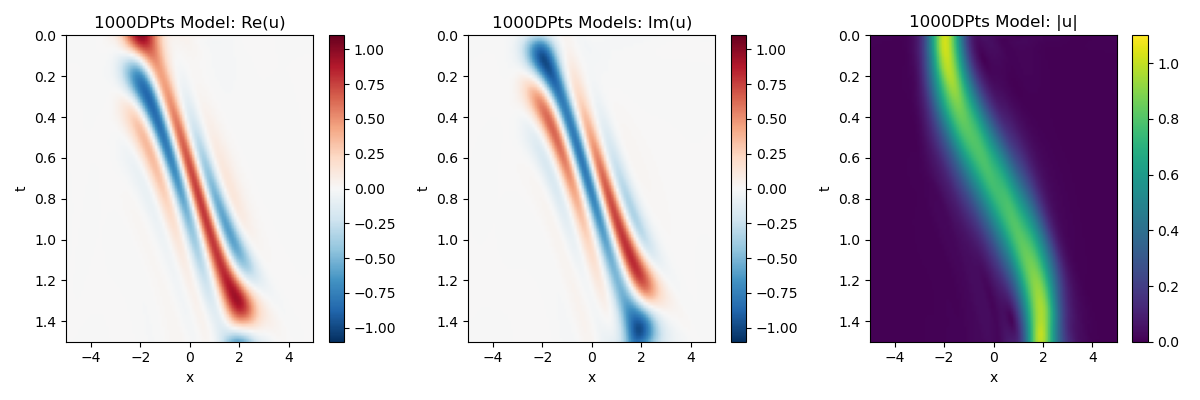}
    \includegraphics[width=0.6\linewidth]{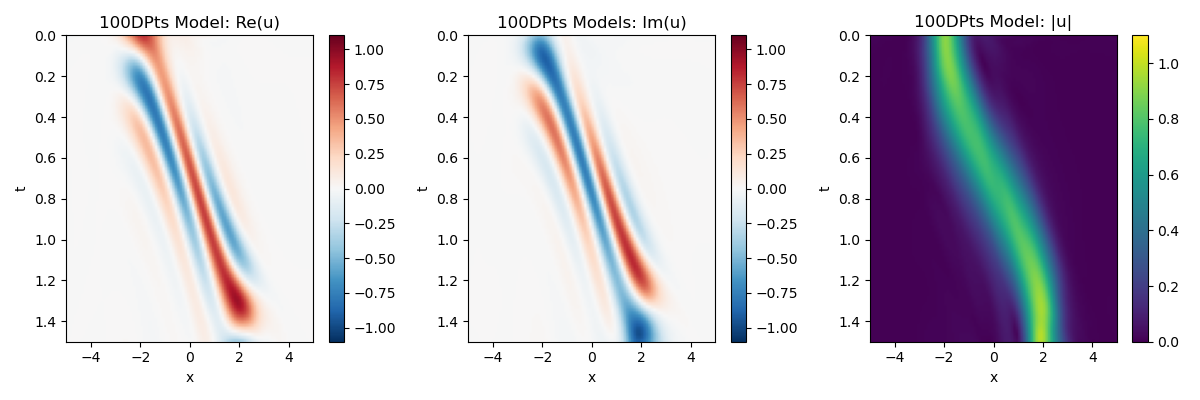}
    \includegraphics[width=0.6\linewidth]{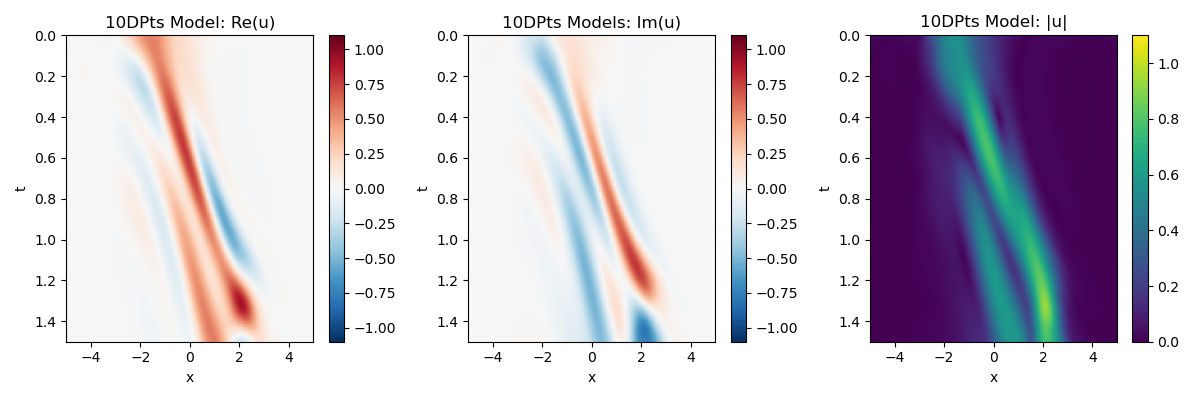}
    \caption{Comparison between numerical solution (top row) and model predictions trained online for the Quantum Harmonic Oscillator. The models were trained with 1000, 100, and 10 data points going from the second to last rows.}
    \label{fig:QHOResults}
\end{figure}

We observe that the method was able to recover the parameters with reasonable precision even when few data points were available. Although the accuracy was reduced with fewer data points, we did not encounter a point of catastrophic failure in our trials. We plot a comparison between solutions trained online for \(k = 4.66\dots\) with varying numbers of data points against the numerical solution generated by FEniCSx in Figure \ref{fig:QHOResults}

We observe that the models trained with 1000 and 100 data points seem to match the shape of the numerical solution to a large extent. We see a sharp decline in accuracy when only training with 10 data points, matching the results from Table \ref{QHO online}.

To compare the performance of IP-Basis PINNs to the standard approach, we implemented the same test on a PINN with an identical architecture to our model. Due to the additional computational costs, we only tested a single set of parameters. To tune the hyperparameters of the PINN, we first tested its performance with \(k=5\) and made adjustments to optimize the absolute error and required training time. The final hyperparameters achieved through this process were 5000 training epochs with the first 1000 only optimizing data and IC losses, an initial learning rate of \(2 \times 10^{-3}\) which is multiplied by  0.9993 at each epoch between epoch 1000 and 3000. The loss weights were identical to the offline training with the exception that \(\omega_{Data}\) remained at a value of 1 for all epochs. Our implementation of PINNs utilized backward mode auto differentiation. The results of this trial are shown in Table \ref{QHO PINN}

\begin{table}[h!]
\centering
\begin{tabular}{| c | c | c | c | c |} 
 \hline
  & Num. Data Points & Final Loss & Absolute Error & Training Time (Seconds) \\ [0.5ex] 
 \hline\hline
 Validation (\(k=5\)) & 10,000 & 5.103E-4 & 2.160E-2 & 424.139\\ 
 \hline\hline
 Test (\(k=2.666\dots\)) & 10,000 & 2.141E-4 & 3.127E-3 & 420.805 \\
 \hline
 Test (\(k=2.666\dots\)) & 1,000 & 3.013E-4 & 4.696E-3 & 424.840 \\
 \hline
 Test (\(k=2.666\dots\)) & 100 & 2.814E-4 & 7.888E-3 & 427.206 \\
 \hline
 Test (\(k=2.666\dots\)) & 10 & 1.688E-3 & 2.529 & 419.835 \\
 \hline
\end{tabular}
\caption{Quantum Harmonic Oscillator standard PINN results.}
\label{QHO PINN}
\end{table}

We can see that in the majority of trials, the standard PINN outperformed IP-Basis PINNs in terms of Absolute Error. This is to be expected, as the framework is designed to produce fast inference times by sacrificing some level of accuracy. In this aspect, our model outperformed standard PINNs by a large margin. Not accounting for the scaling of the framework to larger numbers of readouts (which in the previous sections has been shown to be quite favorable), it performed inference on 10 sets of parameters in 187 seconds in the best-case (10,000 data points) trial shown in Table \ref{QHO online}. The standard PINN required 421 seconds to perform inference for a single parameter with the same number of data points. Thus, if we were to perform inference on the same 10 parameters as done with IP-Basis PINNs, we would expect the standard PINN to take somewhere in the range of 4200 seconds. This corresponds to an approximate 22x improvement in inference time. 

It is also notable that the trial in which the standard PINN performed worse than IP-Basis PINNs was the one that used only 10 data points. While the standard PINN had a catastrophic error in its predicted parameter, the IP-Basis PINN saw a decrease in accuracy but still maintained an error which was an order of magnitude smaller than the standard PINN. We claim that the likely cause of this discrepancy is the inductive bias introduced by the offline training component. In the offline phase, we pre-train the model specifically on quantum harmonic oscillators and in the online phase we simply find a linear transformation from the pretrained network outputs to the solution. It is thus not surprising that the method is able to converge to a sensible solution even with very few data points as the PDE loss landscape should be relatively simple when only training a linear transformation. On the other hand, the loss landscape for the PINN is far more complex. In general, PINNs are quite difficult to train on complex sets of equations such as the quantum harmonic oscillator and often benefit from the presence of training data to assist convergence. When data is unavailable or sparse, it can be extremely challenging to get them to converge. Thus, Basis Networks may be particularly useful when only a small amount of data is available at evaluation time.

\begin{figure}[h!]
\centering
\includegraphics[width=0.55\linewidth]{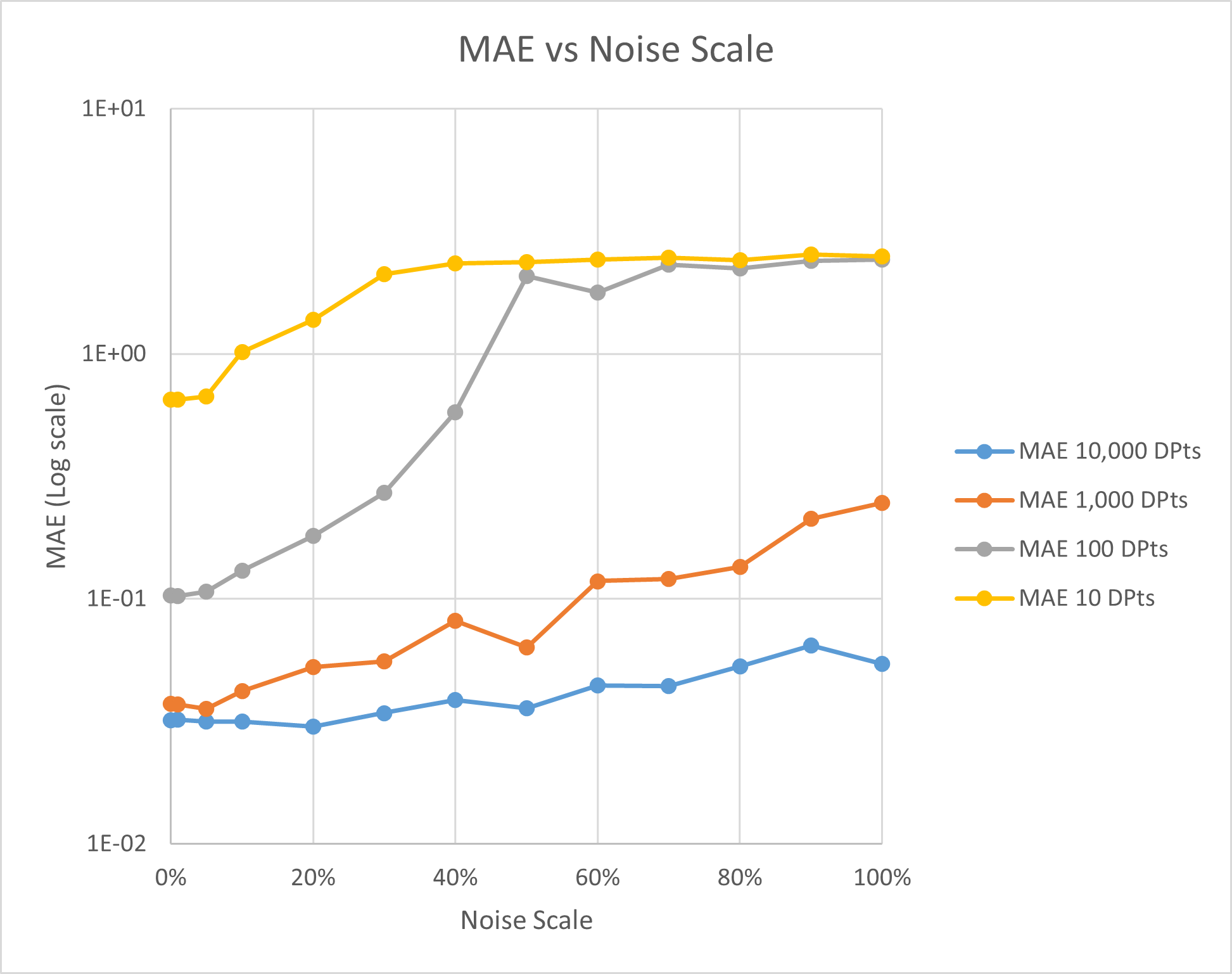}
\caption{\label{fig:QHOMAEvsNoiseScale} Mean Absolute Error of predicted force constant plotted against the scale of injected noise.}
\end{figure}

We then evaluated the model’s performance on noisy input data. Using the pretrained network from the offline phase, we applied the same online training procedure, this time introducing perturbations with magnitudes ranging from 0\% to 100\% of the maximal modulus over the given data. These perturbations were sampled from a uniform distribution scaled to the specified disturbance level. The results of these trials are summarized in Figure \ref{fig:QHOMAEvsNoiseScale}.

We observe that IP-Basis PINNs are capable of making accurate inference for parameters, even when a significant amount of noise is introduced. As would be expected, this robustness to noise improves along with the number of data points available to the method.

\section{Discussion}

In section \ref{results}, we demonstrated the successful application of IP-Basis PINNs to three systems: The damped harmonic oscillator, Lotka-Volterra system, and quantum harmonic oscillator. For each of these systems, we evaluated differing aspects of the method's performance.

For the damped harmonic oscillator, we evaluated the ability of the method to identify unknown coefficients from given data with varying numbers of readouts. We tested models that utilized 10, 30, and 50 readouts during offline training. Surprisingly, the model that performed best for parameters within the offline training distribution was the one trained with only 10 readouts. This begins to change as the initial conditions of the online phase are changed to be beyond those seen in the offline phase. The models trained with more readouts during offline training tended to degrade in performance slower. Even so, we observe that the model trained with 50 readouts performed worse on every metric of parameter accuracy than the model trained with 30 readouts in every trial. we believe that these results can be explained by examining the differences in the final losses of the models in the offline phase. The final loss achieved by the models appears to be directly related to the number of readouts used in training. This trend is likely caused by the increased difficulty in optimizing the more general loss functions that would be required to accurately approximate additional readouts. Allowing the models with more readouts to take more training epochs or be allocated larger models would likely address these problems.

Using the Lotka-Volterra system, we evaluated the performance of the method when used for Universal PINNs \cite{podina_universal_2023}. We again tested the performance of the model on 10, 30, and 50 offline training readouts. We also introduced an additional model trained with 100 readouts. We again see the same trend of increasing loss with a greater number of readouts in the offline phase with a reversal in online performance. while the model with 100 readouts performed worst in offline training, it performed best in the online component. This may be explained by the greater complexity of the Lotka-Volterra system. As a non-linear system of ODEs, learning its dynamics would likely benefit more from additional data points than the damped harmonic oscillator. Thus the increased generalization capabilities of the models with additional readouts may outweigh their reduction in accuracy from training with a more challenging loss function, leading to the observed results. Testing with parameters out of the offline training distribution produced similar results with the model trained with 50 readouts performing slightly better than the one with 100.

Finally, we utilized the quantum harmonic oscillator to test the methods performance on PDEs with varying amounts of data in the online phase, compare with standard PINNs, and evaluate performance with noisy online data. After training the offline phase using 10000 data points, we evaluated online inference given 10000, 1000, 100, and 10 data points. The performance of the model remained relatively consistent when given between 10000 and 100 data points, before seeing a sharp increase between 100 and 10 points. Our comparison with the standard PINN implementation demonstrated the effectiveness of the method for speeding up multi-query inverse problems, showing a 22x improvement in inference time in our trial, and indicated that the inductive bias introduced by the offline training phase may have assisted in converging with only 10 data points where the PINN failed. Our experiments with injected noise demonstrated the robustness of the method for uncertainty in its input.

In \cite{desai_one-shot_2022} the authors derive a closed form expression for the readout layer weights in the special case of linear differential equations, circumventing the need for gradient descent for the forward problem of finding a solution given a set of parameters, initial conditions, and boundary conditions. Unfortunately, their expression currently does not work for inverse problems. This may be a valuable direction for future work. Their work was thus focused entirely on the forward problem. Our work built on theirs by demonstrating the applicability of the method to inverse problems, proposing a method for early stopping, and suggesting the use of forward mode auto differentiation for greater efficiency.

As discussed in Section \ref{introduction}, GPT-PINN is another method for addressing the motivating issues for IP-Basis PINNs \cite{chen_gpt-pinn_2024}. GPT-PINN shares many conceptual similarities to our approach. Both methods utilize an offline-online decomposition, building a surrogate basis for a parametric DE offline, and finding linear combinations of the basis vectors online. With that being said, there are key differences between the methods. Where GPT-PINN enforces that all basis functions are solutions to the DE, the IP-Basis PINNs framework relaxes this restriction, allowing basis functions to be learned freely during training. This difference in restrictions plays a large role in how both methods are trained. 

The offline training routine of GPT-PINNs consists of incrementally adding basis functions chosen according to a greedy algorithm until the overall loss goes under a specified threshold. This approach requires that the basis functions are chosen in series. Since each basis function is a PINN which is trained from scratch, this corresponds to training a series of PINNs one after another.

The architecture of IP-Basis PINNs allows them to take a different approach to training. Instead of sequentially training their basis functions like GPT-PINNs, they optimize a loss function that shapes the basis functions to minimize the loss of many separate PINNs in parallel.

Both methods come with their own costs and benefits. A potential advantage of our approach is parameter efficiency. Architecturally, IP-Basis PINNs promote parameter efficiency by forcing features to be coded by a shared set of parameters, encouraging reuse of similar features between basis functions, whereas GPT-PINNs may learn redundant features across independently trained networks. framework allows features to be coded by a shared set of parameters and reused between basis functions. However, this claim of greater parameter efficiency is a hypothesis based on architectural design; empirical comparative studies are needed to validate it. On the other hand, IP-Basis PINNs are more challenging to grow. While GPT-PINNs can increase their number of basis functions indefinitely until a loss threshold is achieved, the IP-Basis PINNs framework requires a fixed number. If an IP-Basis PINN model is not expressive enough, the likely best option is to train an entirely new one with more parameters and output basis functions. 

The term "basis" implies linear independence, which is not explicitly enforced in our network's training. Future work could analyze the correlations between the outputs of $R$ to determine if they indeed form an approximate basis or a more general spanning set for the solution space.

When choosing which method to use for a specific application, the costs and benefits of both methods should be considered.

\section{Conclusions}
In this work, we examined Inverse-Parameter Basis PINNs (IP-Basis PINNs), a method that utilizes an offline-online decomposition to quickly perform multi-query inference on inverse problems in parametric ODEs and PDEs. The contributions in this work include the novel application of the framework from \cite{desai_one-shot_2022} to inverse problems, the introduction of a validation-based early stopping method for offline training, and the use of forward mode auto-differentiation to significantly improve computational efficiency.

We demonstrated the effectiveness of IP-Basis PINNs in three sets of experiments, examining performance with varying numbers of offline training readouts, parameters outside of the offline distribution, and noisy data in the online phase. We observed that using a greater number of offline training readouts results in a more generally applicable model. However, optimizing these networks becomes more challenging as the number of readouts increases. Initial testing demonstrated that the method may allow for some generalization beyond the parameters and initial conditions observed in offline training. The method was shown to be robust to noise added during the online training phase, with accuracy more strongly related to the number of data points available. This was likely caused by the inductive bias introduced by training the network offline on the clean physics of the problem. Since the basis functions were trained without noise, the network is unable to fit the errors online; it fits the data (to the extent allowed for by the basis functions) and minimizes its residual loss to recover the parameters.

Our results demonstrate that IP-Basis PINNs are a viable and efficient method for producing fast solutions to multi-query inverse problems. The observed significant speedup per query and robustness to sparse and noisy data suggest it should be considered a valuable tool for applications such as real-time inversion and many-query uncertainty quantification. This work positions IP-Basis PINNs as a compelling alternative to other meta-learning approaches like GPT-PINN. The choice between methods will depend on the specific application requirements, weighing factors such as the need for incremental growth against potential gains in parameter efficiency and training simplicity.

Future work will focus on deriving closed-form solutions for the inverse problem where possible, conducting direct empirical comparisons of parameter efficiency with other methods, and further analyzing the properties of the learned basis functions.

\section*{Acknowledgments}
This work was supported by the Natural Sciences and Engineering Research Council of Canada (NSERC). We gratefully acknowledges this support. SM gratefully acknowledges Micheal Alexander for editorial support and insightful discussions on the content.

\nocite{*}
\printbibliography

\end{document}